\documentclass[letterpaper, 10 pt, conference]{ieeeconf}  
\usepackage{graphicx}
\usepackage{diagbox}
\usepackage{multirow}
\usepackage{subcaption}
\usepackage{float}
\usepackage{amsmath}
\graphicspath{{figures/}{pictures/}{images/}{./}} 

\usepackage{tabularx,booktabs}
\newcolumntype{C}{>{\centering\arraybackslash}X} 
\newcolumntype{C}[1]{>{\centering\arraybackslash}p{#1}}
\newcolumntype{P}[1]{>{\raggedright\arraybackslash}p{#1}}
\usepackage{lipsum} 

\IEEEoverridecommandlockouts                              

\title{\LARGE \bf
Benchmarking the Full-Order Model Optimization Based Imitation in the Humanoid Robot Reinforcement Learning Walk
}

\author{Ekaterina Chaikovskaya$^{1}$, Inna Minashina$^{1}$, Vladimir Litvinenko$^{1}$, Egor Davydenko$^{1}$,\\ Dmitry Makarov$^{1, 2}$, Yulia Danik$^{1, 2}$, and Roman Gorbachev$^{1}$
\thanks{$^{1}$Moscow Institute of Physics and Technology
        {\tt\small dorzhieva.em, minashina.ik, davydenko.ev, gorbachev.ra@mipt.ru, litvinenko.vv@phystech.edu}}%
\thanks{$^{2}$Federal Research Center “Computer Science and Control” of Russian Academy of Sciences
      {\tt\small makarov@isa.ru, yuliadanik@gmail.com}}%
}

\begin{document}

\maketitle
\thispagestyle{empty}
\pagestyle{empty}

\begin{abstract}

When a gait of a bipedal robot is developed using deep reinforcement learning, reference trajectories may or may not be used. Each approach has its advantages and disadvantages, and the choice of method is up to the control developer. This paper investigates the effect of reference trajectories on locomotion learning and the resulting gaits. We implemented three gaits of a full-order anthropomorphic robot model with different reward imitation ratios, provided sim-to-sim control policy transfer, and compared the gaits in terms of robustness and energy efficiency. In addition, we conducted a qualitative analysis of the gaits by interviewing people, since our task was to create an appealing and natural gait for a humanoid robot.

According to the results of the experiments, the most successful approach was the one in which the average value of rewards for imitation and adherence to command velocity per episode remained balanced throughout the training. The gait obtained with this method retains naturalness (median of 3.6 according to the user study) compared to the gait trained with imitation only (median of 4.0), while remaining robust close to the gait trained without reference trajectories.

\end{abstract}

\section{Introduction and Related Work }

The problem of locomotion of a bipedal robot is one of the most difficult research topics in robotics, since the anthropomorphic robot is an unstable structure with a large number of degrees of freedom and complex hybrid dynamics. There are two main approaches to designing a walking controller: the first one is based on classical techniques, mostly mathematical optimization, and the other one involves machine learning techniques. Both can be used together and complement each other \cite{doi:10.1177/0278364919859425}. 

In this paper, we focus on the deep reinforcement learning (deep RL) controller design for bipedal locomotion. The deep RL approach is a promising method for bipedal locomotion due to its ability to generalize to different situations and adapt to the robot model and the environment. Disadvantages of this approach include safety concerns due to the inherent stochastic nature of RL algorithms. However, this problem is partially eliminated by specifying high-level constraints on the operation of the motors during the sim-to-real transfer.

\begin{figure}[h]
 \centering
\includegraphics[width=1.\linewidth]{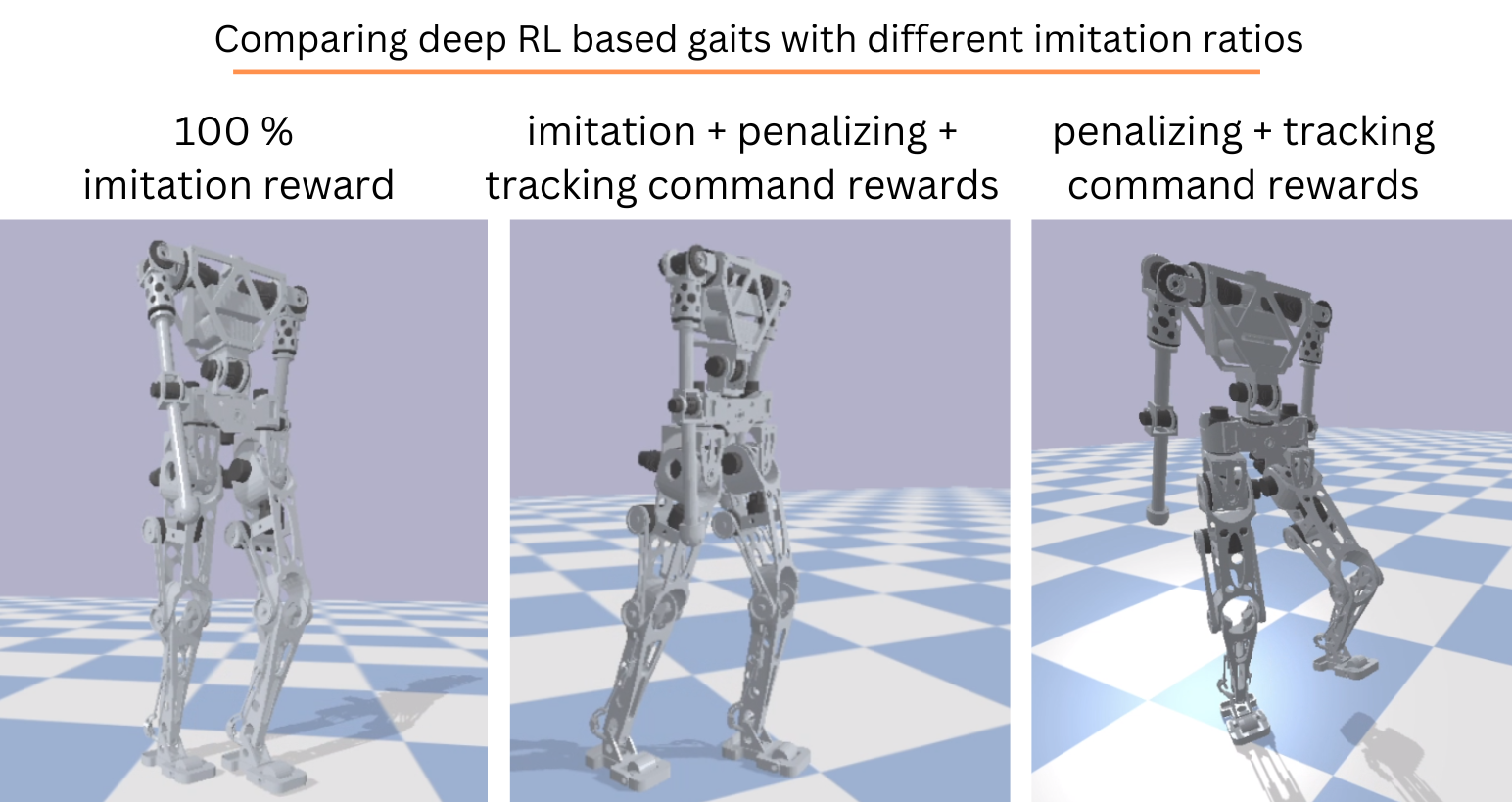}
 \caption{The robot walks in three modes trained by the deep reinforcement learning method with different imitation ratios of reference trajectories for the full-order model.}
 \label{body}
\end{figure}

The deep RL approach is capable of independently exploring the dynamic capabilities of the robot model and generating robust locomotion (the model-free approach). However, this approach has a strong tendency to fall into suboptimal local minima, resulting in visually unattractive and energy inefficient gaits, especially for robots with a high number of degrees of freedom. In addition, the algorithm requires fine-tuning of the training hyperparameters and careful selection of the reward functions. Some forms of the reference trajectory can be added to the learning process to mitigate this. Currently, this methodology is being widely considered for the Cassie series of robots from Agility Robotics \cite{siekmann2020learning}, while in our work we studied the gaits of a humanoid robot with additional degrees of freedom due to an actuated torso and arms. In our work, we investigate the influence of reference trajectories on locomotion learning with the goal of finding a balance between visual appeal and robustness of the final gait.
 

When people talk about an anthropomorphic robot, they usually mean that it will move in a natural way like a human \cite{5354557}. Replicating the trajectory of a human's movement is an intuitive solution, but the morphology of an anthropomorphic robot is different from that of a human, for example, the robot usually has a lower center of mass. Thus, an energy-efficient walking pattern for a human is not optimal for a robot.

If the motion capture data is not used, even the simple reference trajectories in the form of heuristic analysis can improve the overall system performance \cite{10.1109/ICMA54519.2022.9856382}. The more advanced approach is to use some model-based optimization techniques to generate a reference motion. These techniques can use a reduced-order model \cite{9812441} or a full order model \cite{8202230}. Both approaches can be used to augment the RL.

The reference trajectories obtained using simplified models can only capture the most important aspects of complex humanoid model dynamics. The more advanced approach is to use the full-order model, as in Hybrid Zero Dynamics (HZD) \cite{9197175} and Differential Dynamic Programming \cite{6907001}. In this work we used the FROST framework based on the HZD approach \cite{8202230}. With this framework, the user can generate a gait library containing a set of reference trajectories for a range of desired walking speeds using the full-order robot model. This gait library can be used to directly control the robot with appropriate feedback controllers \cite{5160550}. However, using the direct gait library controller is marginally stable and requires an additional heuristic-based controller \cite{8814833, 8796090}.

Depending on the methods used to obtain the reference trajectories, the trajectories may include the positions, velocities, and torques of the motors at each instant and/or the positions and velocities of the feet relative to the center of mass, the center of mass position, the torso orientation, or other data. This data can be used directly in the PD controller as the sum of the reference motor angles and the output of a learned policy, forming a hierarchical control in which the near-optimal legged robot locomotion for the reduced-order model is at the high level and the transfer resistance for the full-order model is at the low level \cite{9380929}. Another possibility is the implicit use of reference trajectories in reinforcement learning rewards. In this case, it is used only during training.

This paper deals with the implementation and qualitative and numerical analysis of full-order model optimization based imitation in the training of the humanoid robot model. Our first contribution is to evaluate the effect of imitation reward on the learning process of bipedal locomotion and to benchmark the resulting robot gaits (Fig. \ref{body}). Our second contribution is the implementation of these gaits on the new full-size anthropomorphic robot model with torso and arms and successful sim-to-sim control policy transfer. In contrast to the robot morphology addressed in \cite{siekmann2020learning}, \cite{9380929}, our humanoid robot model contains 6 degrees of freedom for the legs and 3 degrees of freedom for the fully actuated pelvis. Using the humanoid model with a fully actuated pelvis requires a high quality control policy because the motion of such a model can be directly compared to human walking.


\section{Implementation}
\label{implement}

\begin{figure}[h]
 \centering
\includegraphics[width=0.9\linewidth]{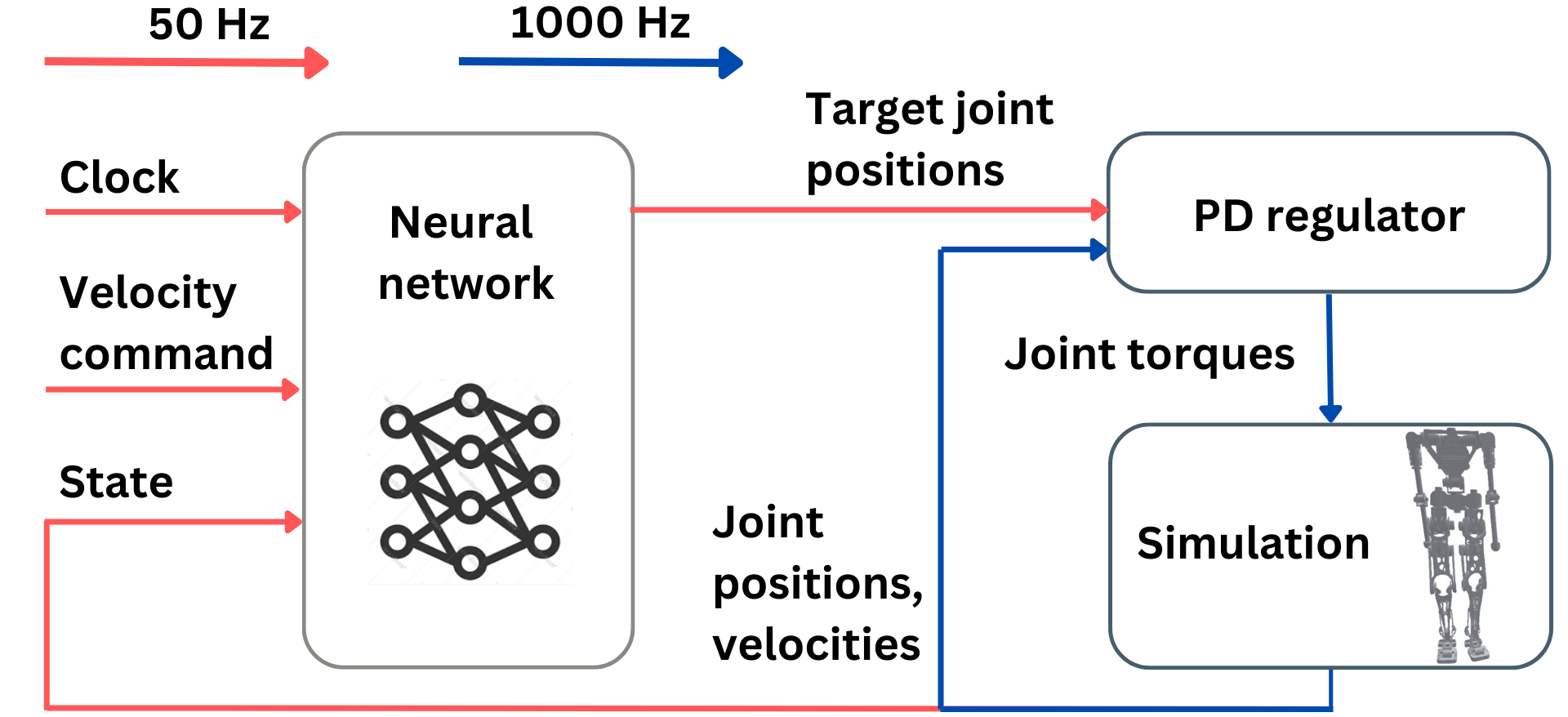}
 \caption{The implementation of the control policy on the full-order anthropomorphic robot model. We used reference trajectories only during training.}
 \label{overview}
\end{figure}

This section presents the implementation of anthropomorphic walking using deep RL augmented with the whole-body motion optimization. We prepared three gaits: in the first one we used only the imitation reward, in the second one, we chose a near-optimal ratio between the imitation reward, the tracking command rewards, and the rewards penalizing inharmonious behavior, and in the third one we did not use the imitation reward at all. The FROST framework was used to obtain the reference trajectories. The policy generated the target joint positions at 50 Hz, then the PD controller calculated the joint torques at 1000 Hz (Fig. \ref{overview}). 

We used a simulation of a humanoid robot model with a height of 1.62 m and a mass of 65 kg, containing 23 degrees of freedom. The robot model was designed by our lab to represent a generic humanoid robot with torque control. Considering the capabilities of the motors that could potentially be part of the robot's construction, we set the joint torque limits (Table \ref{tab_links}).

\begin{table}[h]
\caption{Anthropomorphic robot motor limitations.}
\label{tab_links}
\begin{center}
\scalebox{0.9}{
\begin{tabular}{c c c c}
\toprule
Link name & Max. torque, N*m & Link name & Max. torque, N*m\\
\midrule
Torso yaw & $100$ & Shoulder yaw & $80$ \\

Torso pitch & $125$ & Shoulder pitch & $100$ \\

Torso roll & $100$ & Shoulder roll & $80$\\

Hip yaw & $200$ & Ankle pitch& $200$\\

Hip pitch & $360$ & Ankle roll & $200$\\

Hip roll & $200$ & Elbow pitch & $80$\\

Knee pitch z & $300$ &  & \\
\bottomrule
\end{tabular}
}
\end{center}
\end{table}

\subsection{Simulation Environment}
The deep RL algorithm is based on the enumeration of a large number of possible options for control actions in different states of the robot, followed by the development of a control policy that best allows the robot to perform such actions that lead to an increase in the value of the reward function. Due to the high computational complexity of iterating through a significant number of options, and also due to the fact that reinforcement learning is sensitive to hyperparameters and changes in reward functions, which are often chosen experimentally, an important requirement of the simulator used in the training process is the speed of simulation. In this work, the Isaac Gym simulator \cite{Liang2018GPUAcceleratedRS} and the Legged Gym library, based on the Isaac Gym simulator \cite{Rudin2021LearningTW}, were used to formulate the reward functions and the general structure of the learning process. 

\subsection{Collecting reference trajectories}
The FROST framework was used to obtain the reference trajectories for the full-order robot model. In this framework, motion planning and control generation are based on the use of virtual constraints according to the HZD approach. FROST allows to reduce the task of finding the robot's optimal trajectory to a nonlinear programming problem.

This framework was used to generate a gait library containing the joint trajectories of the agent walking along a horizontal plane in all directions with a maximum lateral velocity of 0.4 m/s and a longitudinal velocity of 1.0 m/s, with a step of 0.1 m/s. 

\subsection{Reward design}
\label{reward design}
The imitation rewards for joint positions and foot positions with respect to CoM given in Eq. (\ref{joint_imit}) and Eq. (\ref{foot_imit}) were used to achieve a natural gait. The rewards given in Table \ref{rew} were used to train the control policies. Their influence on learning is described below. The formulation of the rewards was determined by the works \cite{siekmann2020learning}, \cite{Rudin2021LearningTW}, while their constants were chosen experimentally.

\begin{equation}
 \label{joint_imit}
\centering
 r_{joints} = w * \exp(-\frac{\sum_{i=0}^{N}{||q_i - q^*_i||^2}}{0.2})
, 
\end{equation}
\noindent where $w$ is the weight of the imitation reward,  $q_i$ is the position of the motors, $q^*_i$ is the reference position of the motors for the current command, $N$ is the number of motors.

\begin{equation}
 \label{foot_imit}
\centering
 r_{feet} = w * \exp(-\frac{\sum_{i=0}^{1}{|x_i - x^{*}_i|}}{0.1})
, 
\end{equation}
\noindent where $x_i$ is the current foot position relative to the CoM, $x^{*}_i$ is the reference foot position relative to the CoM for the current command.

\begin{table}[h]
\renewcommand{\arraystretch}{1.2} 
\renewcommand{\tabcolsep}{0.1cm}  
\caption{Reward weights to correct agent behaviour during the learning control policy.}
\label{rew}
\begin{center}
\scalebox{1.0}{
\begin{tabular}{P{2.cm}|C{2.8cm}|C{0.75cm} C{0.75cm} C{0.75cm}}
\toprule
Reward & Equation & Gait 1 & Gait 2 & Gait 3\\
\midrule
Imitation& Eq. \ref{joint_imit} + Eq. \ref{foot_imit} & $5.0$ & $5.0$ & $0.0$ \\

Tracking linear velocity & $\exp(-||V^*_{xy} - V_{xy}|| / \sigma)$ & $0.0$ &$1.5$ & $0.4$ \\

Tracking angular velocity & $\exp(-||\omega^*_{z} - \omega_{z}|| / \sigma)$ & $0.0$ & $1.5$ &$0.3$ \\

No fly & $\sum^{1}_{i=0}{[F_i > 0.1]=1}$ & $0.0$ & $0.1$ & $0.4$ \\

Feet air time & $\sum^{1}_{i=0}{t_i * [F_i<0]}$ & $0.0$ & $3.0$ & $3.0$\\

Action rate & $||a_{t}-a_{t-1}||^2$& $0.0$ & $-3e^{-3}$ & $-5e^{-4}$\\

Motors velocity limit & $\sum^{N}_{i=0}{||\dot{q_i} -\dot{q}_{max}||}$ & $0.0$ & $-0.01$ & $-0.01$\\

Termination & $t < t_{max}$ & $0.0$ & $-30.0$ & $-30.0$\\

Motors positions limit & $\sum^{N}_{i=0}{||q_i -q_{max}||}$ & $0.0$ & $-0.95$ & $-0.95$\\

Orientation & $||g_{xy}||^2$ & $0.0$ &  $-1.0$ & $-1.0$\\

Linear velocity z & $V^2_z$ & $0.0$ & $-1.5$ & $-1.5$\\

Torques & $||\tau||2$ & $0.0$ & $-8e^{-6}$ & $-8e^{-6}$\\
\bottomrule
\end{tabular}
}
\end{center}
\end{table}

\subsection{Implementation details}
\label{learn process}

The anthropomorphic robot was trained to move along a plane with a given commanded linear and angular velocity. The main task was to achieve a visually attractive symmetrical gait similar to human movements, while being resistant to pushes. We used the PPO \cite{https://doi.org/10.48550/arxiv.1707.06347} reinforcement learning algorithm and trained a neural network with 3 fully connected linear layers (512, 256, 128), the training hyperparameters are given in Table \ref{hyperparameters}. 

\begin{table}[h]
\caption{Training hyperparameters.}
\label{hyperparameters}
\begin{center}
\scalebox{1.}{
\begin{tabular}{c c c c}
\toprule
Hyperparameter & Value & Hyperparameter & Value\\
\midrule
Number of agents & $4096$ & Clipping & $0.2$ \\

Number of steps in iteration & $24$ &  Gamma & $0.99$ \\

Entropy coefficient & $0.007$ & Lambda & $0.95$\\

Value function coefficient & $1.0$ & KL target & $0.01$\\
\bottomrule
\end{tabular}
}
\end{center}
\end{table}

The policy was trained to generate the positions of 23 bipedal robot motors: shoulder pitch, roll and yaw, elbow pitch, hip pitch, roll and yaw, knee pitch, ankle pitch and roll for both sides, and torso pitch, roll and yaw. As input, the neural network takes the clock input (a number from 0 to 1), the linear and angular velocities of the center of mass (CoM), the gravity vector, a control command in the form of the CoM target velocity, the current positions and velocities of the robot's joints, and previous actions. The clock input is an important piece of information during training \cite{9561814}. In our case, without clock input, the agents did not learn when trained only with a reward for imitation (gait 1); in other gaits (gaits 2, 3), training was much slower. This is due to the fact that the clock input correlates with the step frequency of the reference trajectories and, together with the command speed, gives a direct indication of which of the reference joint positions should be used.

\subsection{Training using only the imitation reward (gait 1)}
\label{1 mode}

The training of the anthropomorphic robot model included rewards for imitating joint ($r_{joints}$) and foot ($r_{feet}$) positions. The latter was necessary because there was no reward for following the commanded velocity and a motivation was needed to move the agents. First, we trained the policy to follow reference trajectories in the joint space, and then we trained the agent to take the correct position of its body in the environment. As a result, the agent was able to walk with a full body involvement, but in an unstable manner. The characteristics of this gait are discussed in Section \ref{comparison}.

\subsection{Search for the suboptimal balance of rewards (gait 2)}
\label{2 mode}
\begin{figure}[!h!]
 \centering
\includegraphics[width=1.\linewidth]{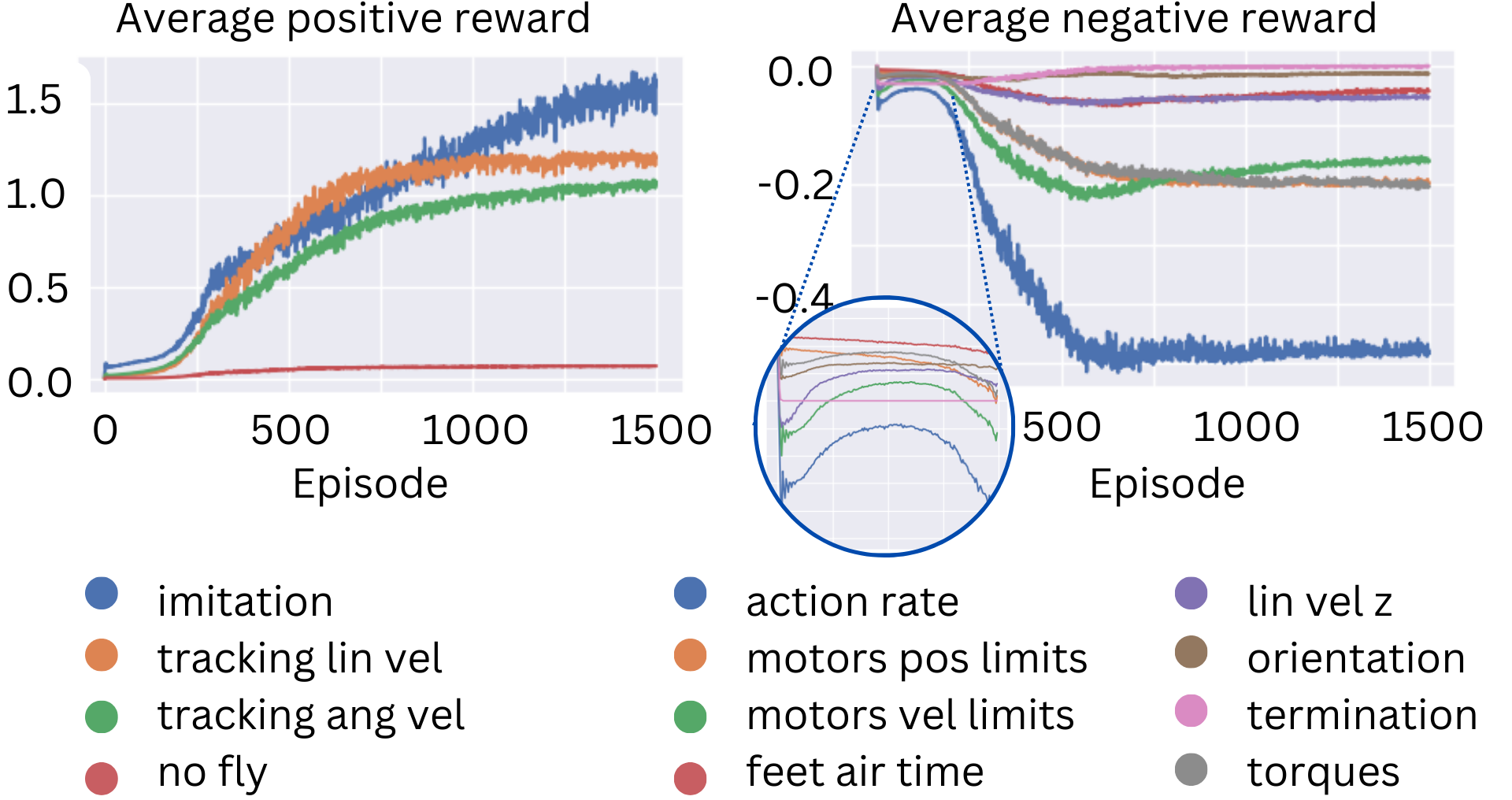}
 \caption{Learning locomotion dynamics of an anthropomorphic robot, gait 2.}
 \label{rew_plot}
\end{figure}

A policy that was both resistant to the application of external pushes and produced a natural gait was achieved by balancing imitation, following the commanded velocity and penalizing rewards. We iteratively adjusted the weights of the positive rewards so that their average values were approximately the same for each episode in this training setup (Fig. \ref{rew_plot}). With a lower value of the imitation reward weight, the resulting gait was unnatural because the agents quickly earned rewards for following command velocities and the reward for imitation did not increase. In another case, increasing the imitation reward resulted in a gait that worked well on a flat surface and did not adapt to perturbations in the form of uneven terrain and pushes, because the agent needed more freedom not to follow the reference trajectory at those moments. For the same reasons we used only $r_{joints}$, without $r_{feet}$ to improve robustness.

The action rate reward had a strong effect on walking because the entropy of the robot is high due to the large number of degrees of freedom, and the reward motivated the agent to be more consistent in its actions. Rewards that made walking energy efficient by limiting the velocity, position, and torque of the motors had about the same effect on learning. At the beginning of training, the termination reward has a large weight that converges to zero as the average episode time increases. The feet air time and no-fly rewards counteract each other by increasing step height without jumping. Because imitation rewards shaped the desired behavior, rewards for trunk movement (orientation and linear velocity (z)) had little effect on the learning.

\subsection{Training without the imitation reward (gait 3)}
\label{3 mode}
We started the process of obtaining a gait with the complete model of the anthropomorphic robot using the reinforcement learning method without reference trajectories. Compared to gait 1 and gait 2, the locomotion is more fractional and jerky. There is also a tendency for non-optimal load distribution between joints: a low center of mass position results in a significant torque at the knees.

\section{Comparison}
\label{comparison}

The three control policies obtained in Section \ref{implement} were compared in terms of command velocity accuracy (Section \ref{track_lin}), robustness (Section \ref{push}), energy efficiency (Section \ref{cost_sec}) and visual appeal (Section \ref{users}). We provided a sim-to-sim transfer to the Pybullet simulator, where all experiments were performed. By transferring to another simulation setup, we wanted to check the stability of the control policy, since we couldn't perform a sim-to-real transfer due to the unavailability of the real robot. 

\subsection{Compliance with linear and angular velocity commands}
\label{track_lin}
\begin{figure}[!h!]
 \centering
\includegraphics[width=0.9\linewidth]{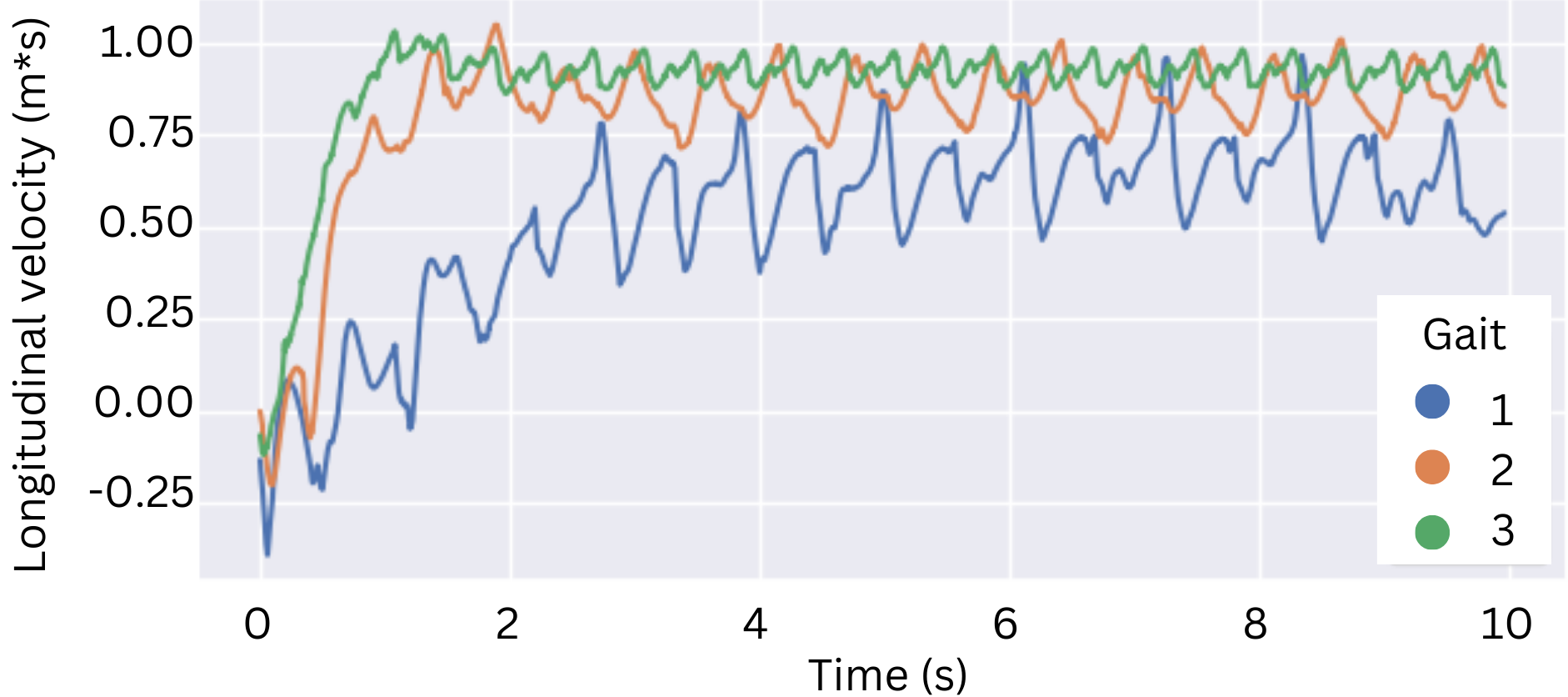}
 \caption{The CoM velocity of the robot, controlled by the obtained policies, at the moment of changing the commanded velocity from 0 to 1 m/s.}
 \label{velocity}
\end{figure}

The next experiment shows the differences in following the commanded velocity. The robot should reach a target velocity of 1 m/s while moving forward from a linear velocity of zero (Fig. \ref{velocity}). A control policy that only used the imitation reward during training (gait 1) is not able to accelerate the full order model of the robot to its maximum speed. In gait 2, the amplitude of the periodic deviations from the target velocity is greater than in gait 3, because the robot's movements are freer and softer due to the imitation reward.

The FROST framework provided reference trajectories for commands in discrete space. However, the final policy was able to interpolate them because we provided commands in continuous space during learning. Second, the FROST framework assumes a two-dimensional command vector containing only linear velocities, without angular velocities. However, the control policy was trained to follow the angular command velocity. As a result, the robot reached a maximum angular velocity of 2.1 rad/s with gait 2 and 3.0 rad/s with gait 3.

\subsection{Push recovery validation}
\label{push}

\begin{figure*}[!h!]
 \centering
\includegraphics[width=0.91\textwidth]{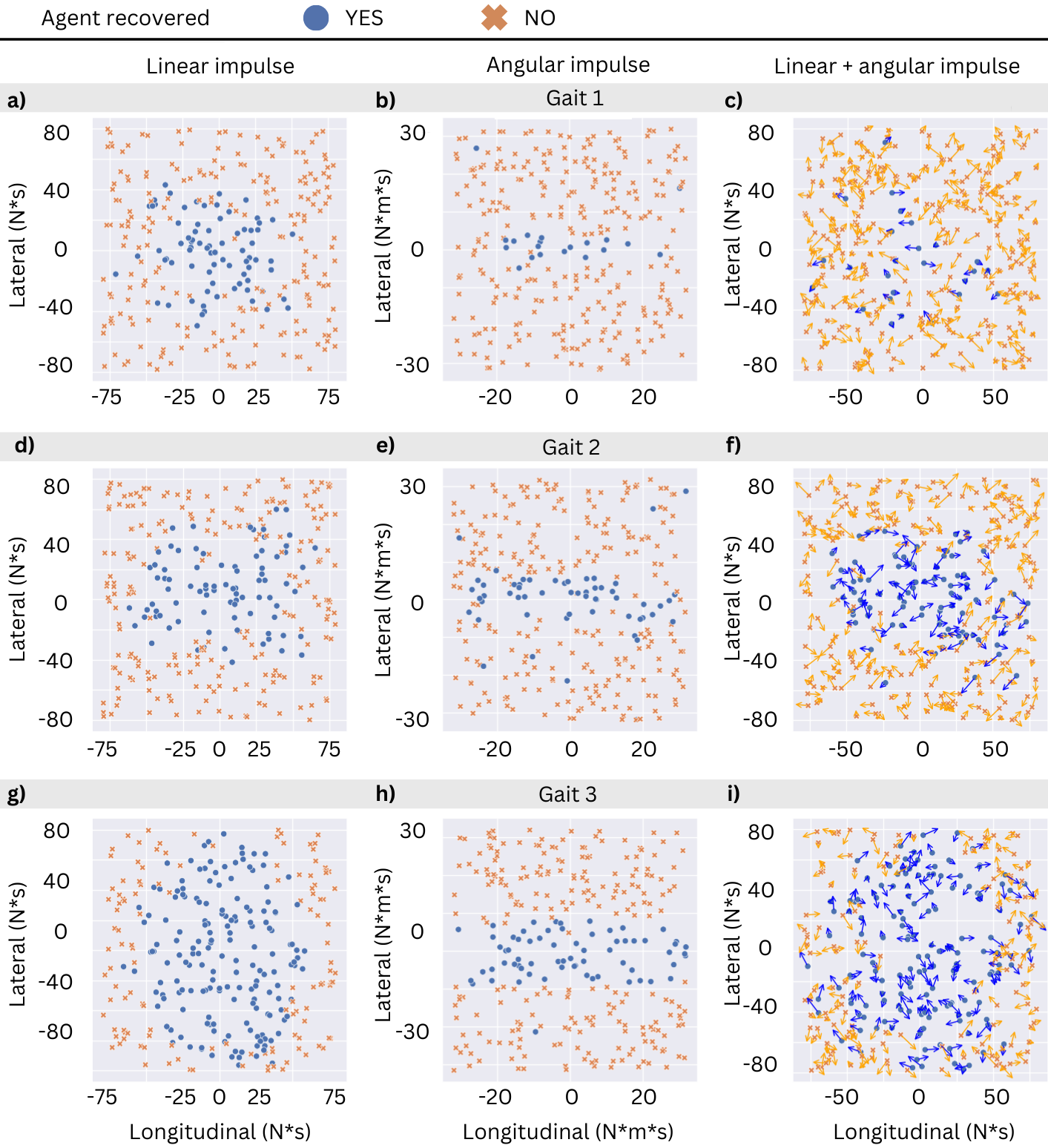}
 \caption{Agent's response to a push under control policy: (a-c). Gait 1, (d-f). Gait 2, (g-i). Gait 3. The first column shows the response to applying a linear impulse to the torso. The second column shows an angular impulse, and the third column shows a linear and an angular impulse to simulate the force applied to different parts of the robot. The percentage of the robot that recovers after the push is: (a) 25$\%$, (b) 7$\%$, (c) 9$\%$, (d) 29$\%$, (e) 19$\%$, (f) 30$\%$, (g) 58$\%$, (h) 22$\%$, (i) 52$\%$.}
 \label{push_recovery}
\end{figure*}

We tested the robustness of the three resulting control policies in the Pybullet environment. External linear and angular impulses were applied to the robot, and falls or successful recoveries from pushes were recorded. For each control policy, a linear impulse was applied to the robot's torso in the first case, an angular impulse in the second case, and both were applied in the third case. Each test contains 300 samples; the robot was given a command velocity of zero during the test. The test results show a significant increase in robustness as the proportion of imitated rewards in training decreases (Fig. \ref{push_recovery}).

\subsection{Cost of transport}
\label{cost_sec}

We measured the energy efficiency of control systems by the cost of transportation \cite{7254255}. To separate the efficiency of the controller and mechanical design from the efficiency of the actuators, two specific metrics can be calculated. One used the total mechanical work of the actuation system $E_m$ (Eq. \ref{em}) and the other included only the positive work $E_{m+}$ (Eq. \ref{em+}). Thus the energetic cost of transport $C_{et}$ and the mechanical cost of transport $C_{mt}$ are given in Eq. \ref{cet}. 

\begin{equation}
 \label{em}
\centering
\begin{split}
 E_m = \int_{0}^{T} \sum_{i=0}^{N}{|\tau_i(t)\omega_i(t)|}\,dt , 
\end{split}
\end{equation}
\begin{equation}
 \label{em+}
\centering
\begin{split}
 E_{m+} = \int_{0}^{T} \sum_{i=0}^{N}{\tau_i(t)\omega_i(t)}\,dt 
 ,\;
     \text{ if } \tau_i(t)\omega_i(t) > 0 
, 
\end{split}
\end{equation}
\noindent where $t$ is the time step, $T$ is the time it takes the robot to walk 1 m, ${N}$ is the number of motors, $\tau_i$ and $\omega_i$ are the motor torque and velocity. 
\begin{equation}
 \label{cet}
\centering
\begin{split}
 C_{et} = \frac{E_m}{mgD}
, \;
 C_{mt} = \frac{E_m+}{mgD}
\end{split}
\end{equation}
\noindent where $m$ is the mass of the robot, $g$ is the gravitational constant, $D$ is the distance traveled by the system ($1m$).

\begin{table}[!h!]
\caption{Results of the cost of transport.}
\label{cost}
\begin{center}
\scalebox{1.}{
\begin{tabular}{c c c c c}
\toprule
Control Scheme & Human & Gait 1 & Gait 2 & Gait 3 \\
\midrule
$C_{et}, J/(kg*m)$ & $0.2$ & $0.59$ & $0.54$ & $0.91$ \\
$C_{mt}, J/(kg*m)$ & $0.05$ & $0.33$ & $0.35$ & $0.62$ \\
\bottomrule
\end{tabular}
}
\end{center}
\end{table}

The cost of transport was calculated as the average of the samples when the robot moves forward at a speed of 0.5 m/s on a flat surface. The results for the cost of transport of the 3 gaits are shown in Table \ref{cost}. A lower value of the metric, close to the human value, is an indicator of an energy efficient gait \cite{9659380}. A higher level of energy efficiency is demonstrated by gaits that include reference trajectories. 


\subsection{Human evaluation of gait performance}
\label{users}
\subsubsection{Participants}
This experiment involved 51 users, twenty-two females and twenty-nine males, aged 16 to 70 years (mean = 33.6, std = 15.3); most of them have never worked with walking robots (39 users), 6 participants have extensive experience, the rest are new to the field; 19 participants have experience in a field related to human anatomy.

\subsubsection{Procedure}
The participants were asked to watch 3 videos with 3 different types of gaits obtained in Section \ref{implement}. They were then asked to rate the natural-looking motion and coordination between the legs and the upper body \cite{7254255} on a five-point Likert scale.

\subsubsection{Human evaluation results}
\begin{figure}[!h!]
 \centering
\includegraphics[width=1.\linewidth]{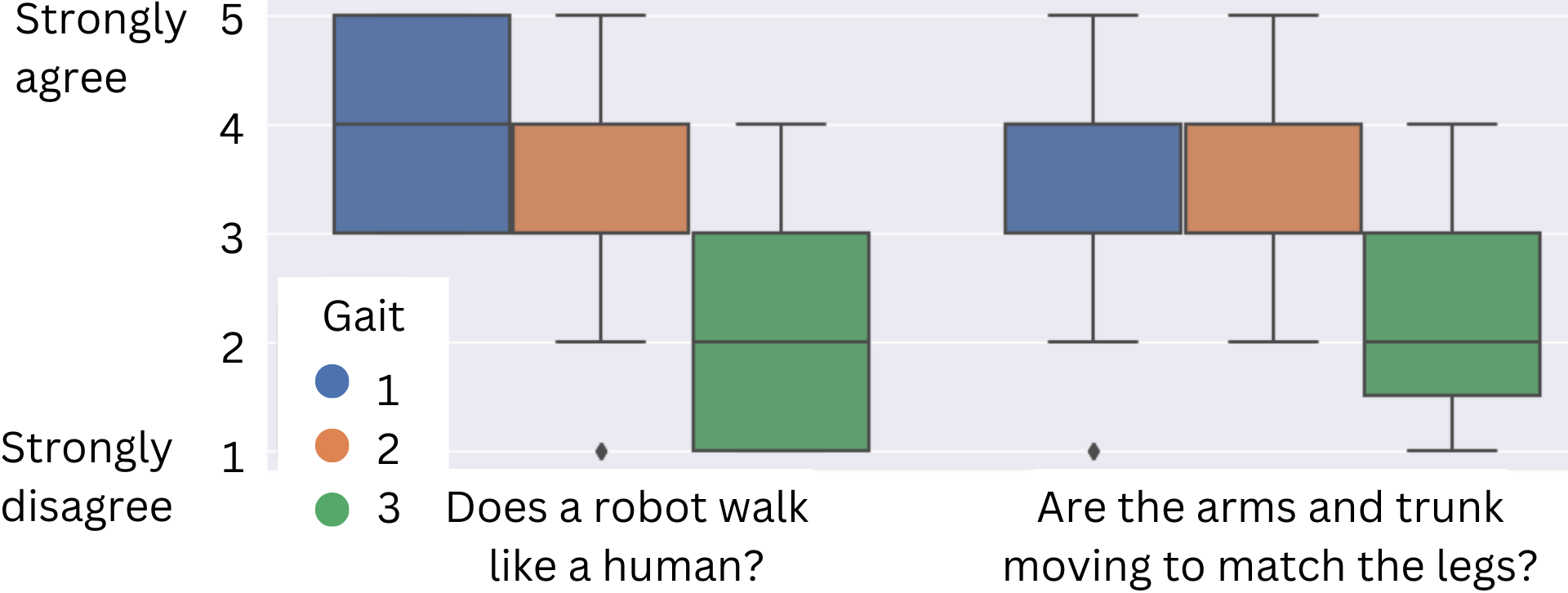}
 \caption{Visual rating of the walking anthropomorphic robot on a 5-point Likert scale based on user responses.}
 \label{us}
\end{figure}

The evaluation (Fig. \ref{us}) confirmed that training with reference trajectories resulted in a more anatomically correct gait (median of 4.0 in gait 1 vs. 3.6 in gait 2 vs. 2.3 in gait 3). Although gait 2 is less natural than gait 1 according to the user study, the metric of coordination between legs and trunk with the arms was found to be equally high for both (median of 3.6 in gait 1 vs. 3.4 in gait 2 vs. 2.2 in gait 3).

\section{Conclusion}

In this paper, three control policies were developed and compared with different ratios of involvement of reference trajectories generated by the full-order model optimization method. Gait 1 was trained using only the imitation reward, gait 2 found an optimal balance between imitation rewards and other rewards that encouraged walking and penalized non-anatomical behavior, and gait 3 did not use imitation.

Gait 2 showed the best energy efficiency, as the control policy improved the metric by rewarding torque reduction, while maintaining the benefits of near-optimal reference trajectories. According to the human evaluation, gait 1 is the most visually appealing, with participants noting the anatomically natural behavior of the robot with the correct movement of the arms and torso. In the test with pushes applied to the robot model, the control policy for gait 3 proved to be more stable than the others. Overall, the control policy for gait 2 remains both natural and robust. In the future, we plan to implement a sim-to-real transfer of gait 2.

\addtolength{\textheight}{-12cm}   





\section*{ACKNOWLEDGMENT}
This work was supported by a grant for research centers in the field of artificial intelligence, provided by the Analytical Center for the Government of the Russian
 Federation in accordance with the subsidy agreement (agreement identifier 000000D730321P5Q0002) and the agreement with the Moscow Institute of Physics and Technology dated November 1, 2021 No. 70-2021-00138.

\bibliographystyle{IEEEtran}
\bibliography{root}

\end{document}